\documentclass[runningheads]{llncs}
\usepackage[T1]{fontenc}
\usepackage{graphicx}
\usepackage{booktabs}
\usepackage[misc]{ifsym}

\usepackage{amsmath} 
\usepackage{amssymb} 
\usepackage{todonotes}
\usepackage[bottom]{footmisc}

\newcommand{\NN}{\mathbb{N}}

\usepackage{subcaption}
\usepackage{graphicx}%
\usepackage{wasysym}%
\DeclareMathOperator{\Azero}{\text{\female}}
\DeclareMathOperator{\Aone}{\text{\male}}

\DeclareMathOperator{\groups}{\{\Azero,\Aone\}}

\begin{document}

\title{``Patriarchy Hurts Men Too.''\\ Does Your Model Agree?\\ A Discussion on Fairness Assumptions}

\titlerunning{A Discussion on Fairness Assumptions}


\author{Marco Favier \and Toon Calders}



\institute{University of Antwerp, Antwerp, Belgium \email{\{marco.favier, toon.calders\}@uantwerpen.be}}

\maketitle

\begin{abstract}
The pipeline of a fair ML practitioner is generally divided into three phases: 1) Selecting a fairness measure. 2) Choosing a model that minimizes this measure. 3) Maximizing the model's performance on the data. In the context of group fairness, this approach often obscures implicit assumptions about how bias is introduced into the data. For instance, in binary classification, it is often assumed that the best model, with equal fairness, is the one with better performance. However, this belief already imposes specific properties on the process that introduced bias. More precisely, we are already assuming that the biasing process is a monotonic function of the fair scores, dependent solely on the sensitive attribute.
We formally prove this claim regarding several implicit fairness assumptions. This leads, in our view, to two possible conclusions: either the behavior of the biasing process is more complex than mere monotonicity, which means we need to identify and reject our implicit assumptions in order to develop models capable of tackling more complex situations; or the bias introduced in the data behaves predictably, implying that many of the developed models are superfluous.
\end{abstract}
\section{Introduction}
It is now well known that AI models may exhibit unfair behaviors against historically marginalized groups and minorities due to biases present in the data used to train these models. 
Famous cases like \emph{SyRI}~\cite{appelman2021social}, the \emph{toeslagenaffaire}~\cite{hadwick2021lessons}, and the \emph{COMPAS}~\cite{angwin2022machine} algorithm have proven beyond any doubt that the risks involved in using classifiers without concern for bias are real and have negatively affected, and still affect, communities that already suffer from unjust discrimination.

Since this acknowledgment, much effort has been spent in the last decade to study the phenomenon from a scientific and philosophical perspective, and many solutions have been proposed to improve the fairness of models.

Group fairness measures try to tackle this problem by evaluating metrics that describe how unfairly entire sensitive groups have been treated compared to each other.
For instance, the \emph{Demographic Parity Difference} evaluates how statistically likely it is to be assigned a positive label for one group against another. 
Similarly, the \emph{Equal Opportunity Difference} computes the difference between the false negative rates evaluated for different communities. Barocas et al.~\cite{fairnessbook} provide a systematic overview of the most common fairness measures.

On top of these measures, many different models able to minimize (un)fairness measure have been developed.
In general, these models are evaluated by checking which one performs best according to fairness and/or accuracy, with the underlying assumption that, as long as the fairness measure is low, the preferable model is the one that performs better.

Nonetheless, understanding when or even why a model is better than another is not an easy question to answer; fairness will always remain undoubtedly a subjective concept, and any comparison between models or fairness measures is reasonable only within very specific contexts.

But also formally, the field of fairness lacks a theoretical and foundational background to better explain fair models and their behavior. It is often unclear how the choice of a specific model impacts the outcome and under what circumstances a given model should be used.

We believe that one of the reasons for this confusion comes from the fact that a number of these models seem to work under similar assumptions, trying to be as faithful as possible to the data. But, if we believe that our data is biased and thus not representative of the population, it's debatable if fidelity is the best approach to fairness.

This assumption is often implicitly made and not discussed, and it has important practical but also theoretical consequences on how we should approach fairness.

In this work, we explore a number of implicit assumptions often present in fairness measures and models.
In particular, we show how these assumptions all share the same underlying idea: the biasing process, that is, the process that has introduced bias into our data, is a monotonic function on each sensitive group. 
This means that if two individuals from the same sensitive group are ranked fairly, then the one with the highest score will always be ranked higher according to unfair probabilities as well. People are swapped in the ranking only if they belong to different sensitive groups.

But this has very strong implications and often does not reflect reality. For instance, consider a model screening CVs for a job position that unfairly privileges men. If a male candidate writes using the first person, while another uses the third person, the model may privilege the latter.
A different outcome only because it's easier to assess the gender from all the ``he"s instead of the ``I"s. Wouldn't it then be fair to improve the ranking of the first candidate as well?

Discrimination within the same sensitive group is a well-known phenomenon, so discussing whether a model should consider it or not is an important question that shouldn't be implicitly answered.

So, is your model aware that the patriarchy hurts men too?
\section{Related Work}
We work under the so-called \emph{Fair World Framework}~\cite{favier2023fair}. Under this framework, it is assumed that an inaccessible fair world exists. This would be an ideal world where no bias or prejudice exists, untouched by the systemic issues our world suffers from. Data sampled from this world would be considered fair, and decisions based on it would be considered unbiased. We don't have access to this world; instead, we have access to data collected from our world, which is a biased version of the fair world.

This framework has been used to study the relationship between fairness and accuracy \cite{wick2019unlocking,friedler2021possibility,favier2023fair}, providing a useful tool to understand the objective of fairness in ML. It represents a paradigm shift from the classical fairness definition: a fair model is not one that satisfies some fairness constraint, but one that is accurate on the fair world.

We will adopt this framework to explore the space of theoretically optimal fair models for multi-label classification. For binary classification, Corbett-Davies et al.~\cite{Corbett-Davies-faircost} and Menon et al.~\cite{menon-faircost} have proven that the best fair model, where fairness means minimizing a fairness measure, can always be achieved by applying a different threshold for each sensitive group of the score function. Our work will explore how their results apply within the fair world framework.

On the other hand, Wick et al.~\cite{wick2019unlocking} have shown that a fair model can be more accurate than an unfair one. They argue that when data are biased, fairness can improve accuracy on the originally unbiased data, contrary to the common belief that fairness and accuracy are in trade-off.

They claim that this phenomenon is often misunderstood because we do not have access to unbiased data, and we should not simply assume that the data at hand is the best data to evaluate a model.

More recently, Goethals et al.\cite{goethals2024beyond} have conducted experiments to see if fair models change the ranking within sensitive groups. Our framework provides a much-needed theoretical background for their results. Moreover, analyzing how a fair model can impact individuals within the same sensitive group has also been explored in the context of individual fairness~\cite{dwork2012fairness,individual_fairness_speicher}.

\section{Notation and Problem Setting}
We will focus on a fairness multi-label classification problem. More formally, we consider a probability space $(X,\mu)$ called the \emph{feature space}, where
\begin{itemize}
    \item $X$ is the set of possible individuals, each described by a feature vector $x\in X$.
    \item $\mu$ is a probability measure used to sample individuals from  $X$.
\end{itemize}
In a classical group fairness fashion, we assume that the set $X$ is partitioned into different sensitive groups. 
These groups generally represent communities that have historically been treated differently for a variety of unjust reasons. For instance, the sensitive groups can represent different genders or ethnicities. 
For convenience, we consider only two sensitive groups. More formally, a \emph{sensitive partition} $\{\Azero,\Aone\}$ of $X$ is a binary partition of $X$ into non-zero measurable sets. An element $A\in\{\Azero,\Aone\}$ is called a \emph{sensitive group}.

The reader should be aware that all our results generalize to any number of sensitive groups. Moreover, the symbols ``$\Azero$'' and ``$\Aone$'' used to identify the groups are completely arbitrary; they are not meant to represent genders or any other specific sensitive attribute.

Together with the space $X$, we also have a \emph{categorical label} $Y\in\{y_1,\dots, y_n\}$  that is assigned to each individual. The main difference between a classical decision problem and our setting is how the label is sampled for each individual.

Instead of assuming the existence of a single conditional distribution, we consider two of them: $p_u(\boldsymbol{y}\,\vert x)$ and $p_f(\boldsymbol{y}\,\vert x)$. More specifically:
\begin{itemize}
    \item $p_u(\boldsymbol{y}\,\vert x):=\big(p_u(y_1\vert x),\dots,p_u(y_n\vert x)\big)$ is the observable distribution that we can infer from the data we have, but it is considered unfair towards some sensitive groups due to its biased information.
    \item $p_f(\boldsymbol{y}\,\vert x):=\big(p_f(y_1\vert x),\dots,p_f(y_n\vert x)\big)$ is instead the fair distribution. Decisions based on this distribution are considered unbiased, but it's not possible to access it directly. This distribution represents what we would observe if there were no bias or prejudice in our society.
\end{itemize}
As conditional distributions, both $p_f$ and $p_u$ are measurable functions
\[
p_u,p_f\colon X \to \Delta^{n}
\]
where $\Delta^{n}$ is the $n$-dimensional simplex: 
\[
    \Delta^{n}:=\{\mathbf{t}\in [0,1]^n\colon t_1+\dots+t_n=1\}
\]

The relationship between the fair and unfair distributions is determined by a \emph{biasing process} $\beta$ that transforms fair probabilities into unfair ones. Similarly, we can describe a \emph{debiasing process} $\beta^{-1}$ that does the opposite.
We can show that these processes always exist:
\begin{lemma}
    Given probability distributions $p_f$ and $p_u$, there exists a biasing and a debiasing process, $\beta,\beta^{-1}\colon \Delta^n\times X\to \Delta^n,$ respectively, such that
    \[
        p_u(\boldsymbol{y}\,\vert x) = \beta\left(p_f(\boldsymbol{y}\,\vert x), x\right)\text{ and } p_f(\boldsymbol{y}\,\vert x) = \beta^{-1}\left(p_u(\boldsymbol{y}\,\vert x), x\right)
    \]
    \begin{proof}
        The proof is almost tautological: since $\beta$ and $\beta^{-1}$ depend directly on X, then the use of the conditional distributions is superfluous. We can simply set 
        \[
            \beta\left(- , x\right):=p_u(\boldsymbol{y}\,\vert x) = \text{ and } \beta^{-1}\left(-, x\right):=p_f(\boldsymbol{y}\,\vert x)
        \]
        which proves the thesis.
    \qed\end{proof}
\end{lemma}
Formally, using $\beta^{-1}$ to indicate the debiasing process is an abuse of notation, since the debiasing process is not technically the inverse of the biasing process. Nonetheless, we will use this notation to underline their mutual relationship.

The fact that the biasing (or debiasing) process depends on $x$ as well as $p_f$ (or $p_u$) may be confusing. This is done to underline that the biasing process transforms the fair probabilities into unfair ones. However, this requirement alone is not sufficient, as the transformation may also vary depending on the specific individual considered, necessitating the dependency on $x$.

Our objective will be to assign a label to each individual using a \emph{decision function}. Formally, a decision $\boldsymbol{d}$ is a function 
\[
    \boldsymbol{d}\colon X\to \Delta^n\quad x\mapsto \boldsymbol{d}(x)=\left(d_1(x),\dots,d_n(x)\right)
\]
The use of $\Delta^n$ as the codomain of a decision allows us to consider probabilistic decisions: for a data point $x$, $\boldsymbol{d}(x)$ represents a distribution on $Y$, from which we sample a predicted label $\widehat{Y}\in \{y_1,\dots, y_n\}$. Notice that this doesn't force us to consider only probabilistic decisions; in fact, we can also consider deterministic decisions by simply setting $d_i(x)=1$ for some $i\in[n]$.

We can evaluate a decision based on how well it performs on each label. In particular, we define  $P(\widehat{Y} = y_i, Y = y_j)$ as the following value:
\[
    P(\widehat{Y} = y_i, Y = y_j):=\int_{X}d_i(x)p(y_j\,\vert x)\mu(x)
\]
We will use $P_u$ when the calculation is computed using $p_u(\boldsymbol{y}\,\vert x)$, and similarly $P_f$ when fair probabilities are used. This allows us to define the following:
\begin{definition}[Pareto Order]
    Given two decisions $\boldsymbol{d},\boldsymbol{d}'$, we will say that $\boldsymbol{d}$ is Pareto \emph{non-inferior} (or \emph{non-superior}) to $\boldsymbol{d}'$, and write $\boldsymbol{d}\geq \boldsymbol{d}'$ if and only if for all $i\in [n]$
    \[
        P(\widehat{Y}= y_i, Y=y_i)\geq P(\widehat{Y}' = y_i, Y=y_i)
    \]
where $\widehat{Y}\sim \boldsymbol{d}(x)$ and $\widehat{Y}'\sim \boldsymbol{d}'(x)$. We will use $\geq_u$ or $\geq_f$ to indicate according to which conditional probability $p_f,p_u$ we have calculated $P(\widehat{Y}= y_i, Y=y_i)$.
\end{definition}
It's important to notice that if $\boldsymbol{d}\leq\boldsymbol{d}'$ according to some probability, $\boldsymbol{d}'$ is better in the sense that it can better represent the data. It doesn't mean that $\boldsymbol{d}'$ is more fair.

The Pareto order on decisions is not a total order, meaning two decisions may not be comparable. So it is possible that $\boldsymbol{d}\not\geq \boldsymbol{d}'$ and $\boldsymbol{d}'\not\geq \boldsymbol{d}$. Moreover, there are cases where two different decisions $\boldsymbol{d}\neq \boldsymbol{d}'$ are equivalent, meaning $\boldsymbol{d}\geq \boldsymbol{d}'$ and $\boldsymbol{d}'\geq \boldsymbol{d}$. 

We are obviously interested in finding the best fair decision possible, that is, a decision that is Pareto maximal according to $\leq_f$.

\begin{definition}[Unbiased Decision]
    A decision $\boldsymbol{d}$ on $U\subseteq X$ is said to be unbiased if it is maximal on $U$ according to $\leq_f$ and depends only on fair probabilities. Formally:
    \begin{itemize}
        \item $\boldsymbol{d}$ is Pareto maximal for $\leq_f$; that is, if $\boldsymbol{d}'\geq_f \boldsymbol{d}$ then $\boldsymbol{d}\geq_f \boldsymbol{d}'$ also holds.
        \item $\boldsymbol{d}$ depends only on fair probabilities; that is, there exists a function $\boldsymbol{D}$ such that $\boldsymbol{d}(x) = \boldsymbol{D}(p_f(\boldsymbol{y}\,\vert x))$.
    \end{itemize}
\end{definition}
Notice that, since we do not have access to the fair distribution $p_f$, we are unable to determine if a decision is unbiased or not. This is why we need to use fairness measures as proxies to ensure fair decisions. The following example should clarify this point.
\begin{example}
    Suppose every individual in $X$ is described by two independent features: the number of languages they speak, ranging from one to three ($s={1,2,3}$), and whether they are religious or not ($r={0,1}$), which is a sensitive attribute.
    Moreover, all six possible combinations of attributes appear to be equally distributed. You want to predict whether a person will obtain a PhD in their lifetime ($y=\{0,1\}$). The fair probabilities are the following:
    \begin{align*}
        p_f(y=1\vert s,r=0)=
        \begin{cases}
            1\% & \text{if } s=1\\
            2\% & \text{if } s=2\\
            4\% & \text{if } s=3
        \end{cases}
        & 
        &p_f(y=1\vert s,r=1)=
        \begin{cases}
            1\% & \text{if } s=1\\
            3\% & \text{if } s=2\\
            3\% & \text{if } s=3
        \end{cases}
    \end{align*}
    These probabilities can be considered fair according to demographic parity: the chances of obtaining a PhD are the same for religious and non-religious individuals.\\ 
    Sadly, academia is known to be biased against religion. Consequently, in our unfair society, many religious people don't even apply for a PhD. As a result, the unfair probabilities are as follows:
    \begin{align*}
        p_u(y=1\vert s,r=0)=
        \begin{cases}
            1\% & \text{if } s=1\\
            2\% & \text{if } s=2\\
            4\% & \text{if } s=3
        \end{cases}
        & 
        &p_u(y=1\vert s,r=1)=
        \begin{cases}
            0\% & \text{if } s=1\\
            1\% & \text{if } s=2\\
            2\% & \text{if } s=3
        \end{cases}
    \end{align*}
    Notice how the new probabilities do not satisfy demographic parity anymore. Consider now the decision $\boldsymbol{d}$ such that
    \[
        d_1(s,r)=
        \begin{cases}
            1 & \text{if } p_u(y=1\vert s,r)\geq 2\%\\
            0 & \text{otherwise}
        \end{cases}
    \]
    Despite being purely based on the probabilities derived from the data, the decision is biased because $d_1(2,1)=0$ and $d_1(2,0)=1$, whereas $p_f(y=1\vert s=2,r=1)> p_f(y=1\vert s=2,r =0)$.
    Instead, if we take the decision $\boldsymbol{d}'$ such that
    \[
        d'_1(s,r)=
        \begin{cases}
            1 & \text{if } p_u(y=1\vert s,r)\geq 2\% \text{ and } r=0\\
            1 & \text{if } p_u(y=1\vert s,r)\geq 1\% \text{ and } r=1\\
            0 & \text{otherwise}
        \end{cases}
    \]
    then $\boldsymbol{d}'$ is unbiased since it's equivalent to 
    \[
        d'_1(s,r)= 1 \iff p_f(y=1\vert s,r)\geq 2\%
    \]
    Notice how the second decision explicitly depends on the sensitive attribute $r$, which is necessary to contrast the bias present in the data.
\end{example}
Unbiased decisions are not always able to satisfy fairness constraints. For instance, in the previous example, it's not hard to find unbiased decisions that are still not fair according to demographic parity.

But it is true that if there are differences in the outcomes for the two communities, these can be fully explained using fair probabilities. This framework allows us to discuss fairness even in cases where the communities have different distributions. This is because the definition of fairness is now tied to the fair world and its probabilities. For instance, if we assume that $p_f$ is identically distributed on $\Azero$ and $\Aone$, then unbiased decisions will also be fair.

This idea formalizes why fairness metrics and losses can be a useful tool for finding fair and unbiased decisions: within our framework, including a penalty term for unfairness in a loss function can be viewed as placing a prior on the decision space that favors unbiased decisions.

That being said, finding optimal decisions, especially in multi-label classification, is not an easy task. Fortunately, Weller's Theorem~\cite{weller1985fair} provides a way to find such decisions when we are able to model the underlying probability distribution. To introduce it we need first the following definition:
\begin{definition}
	A sequence $(\boldsymbol{\omega}^{(k)})_{k\in\NN}$ of point in $\Delta^n\cap (0,1)^n$ is said to be \emph{well-behaved} if for all $i,j\in [n]$ the limit $\lim_{k\to \infty} {\omega_i^{(k)}}/{\omega_j^{(k)}}$
	converges in $[0,\infty]$.
\end{definition}
\begin{theorem}[Weller's Theorem \cite{weller1985fair}]\label{teo:weller}
    A decision $\boldsymbol{d}\colon X \to \Delta^n$ is Pareto maximal if and only if there exists a well behaved sequence $(\boldsymbol{\omega}^{(k)})_{k\in\NN}$ such that almost everywhere on $X $, for all $i,j\in [n]$
    \[
        d_i(x)>0 \Rightarrow \exists\ \overline{k}\in\NN\text{ s.t. }\frac{p(y_i\vert x)}{p(y_j\vert  x)}\geq \frac{\omega_i^{(k)}}{\omega_j^{(k)}} \quad \forall k\geq \overline{k}
    \]
    Such sequence $(\boldsymbol{\omega}^{(k)})_{k\in\NN}$ is said to be $w$-associated with $\boldsymbol{d}(x)$.
\end{theorem}
\begin{figure*}[t!]
	\centering
	\begin{subfigure}[t]{.32\textwidth}
		\centering
		\includegraphics[width=\textwidth]{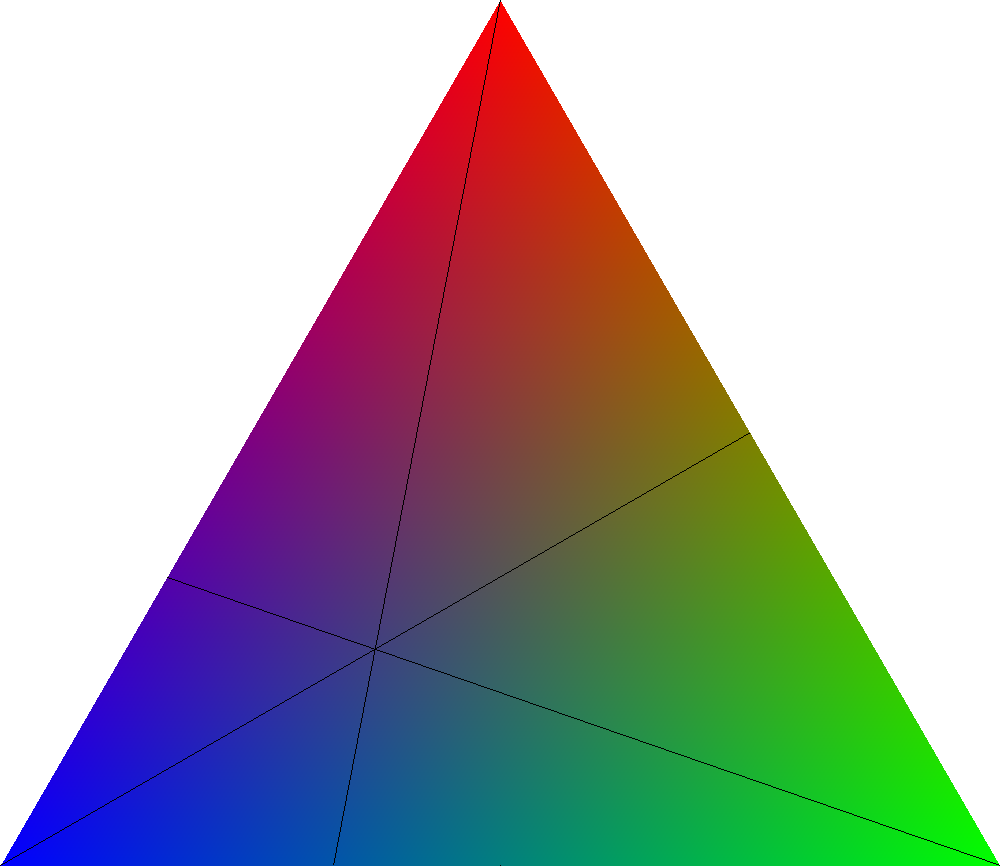}
		\caption
		{
			Lines intersect at point $(0.25,0.25,0.5)$
		}
	\end{subfigure}%
	\hfill
	\begin{subfigure}[t]{.32\textwidth}
		\centering
		\includegraphics[width=\textwidth]{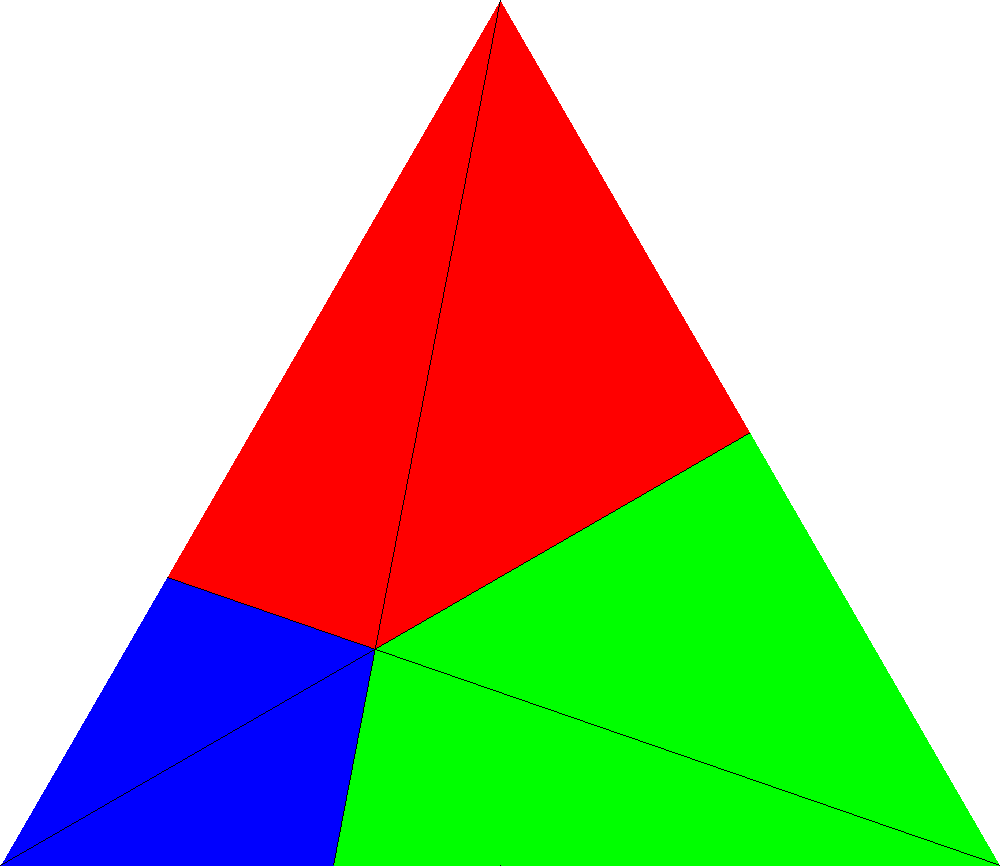}
		\caption
		{
			$\boldsymbol{\omega}^{(k)}=\left(\dfrac{1}{4},\dfrac{1}{4}, \dfrac{1}{2}\right)$
		}
	\end{subfigure}
	\hfill
	\begin{subfigure}[t]{.32\textwidth}
		\centering
		\includegraphics[width=\textwidth]{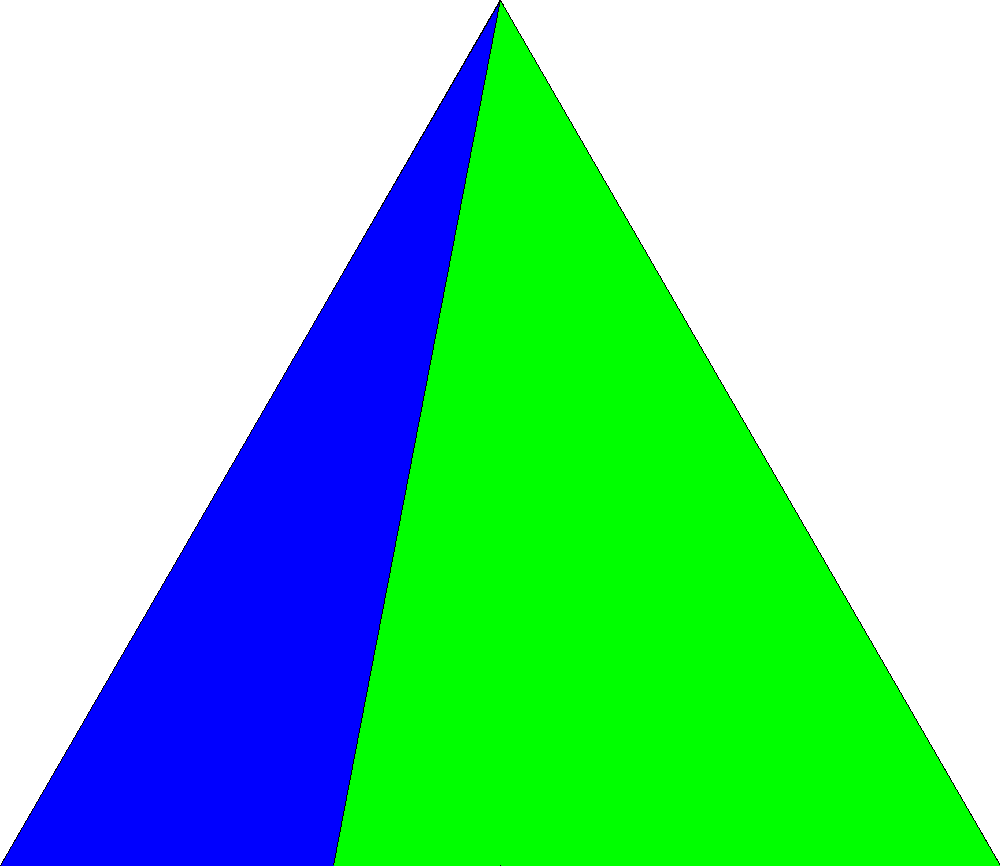}
		\caption
		{
			$\boldsymbol{\omega}^{(k)}=\left(\dfrac{k-9}{k},\dfrac{3}{k}, \dfrac{6}{k}\right)$
		}
	\end{subfigure}
	\caption{Optimal decisions according to Weller's Theorem on the $\Delta^3$ simplex with labels $Y=\{\text{red},\text{green}, \text{blue} \}$. The RGB decomposition of a color represents the conditional distribution $p(\boldsymbol{y}\vert x)$.}
\end{figure*}

\section{The Group Fairness Assumptions}
\label{sec:assumptions}
In this section, we will introduce a number of assumptions that fairness literature often implicitly uses. 
Before continuing, we want to make it clear that we don't think having assumptions is wrong, nor that these specific assumptions necessarily are. In fact, we believe that having assumptions is often unavoidable when developing a model. The famous quote by George Box, ``All models are wrong, but some are useful,'' makes it explicit that without assumptions on our data, even overly simplistic ones, we couldn't build a model. Even the entirety of the field of fair ML is based on the assumption that some sensitive attributes should not influence decisions.

It is important, though, to make these assumptions explicit, especially when they are used to justify the development of a model. In the same way, it is also important to realize the consequences of these assumptions. Fair ML is a clear example of the importance of doing so: when researchers began formalizing the assumption that a model should not be biased, they realized there were different ways to achieve this, to the point that different views on fairness can conflict with each other. Without stating the assumptions on fairness, we wouldn't have been able to prove the now-famous impossibility results \cite{friedler2021possibility,kleinberg2016inherent,saravanakumar2020impossibility,miconi2017impossibility}.

Moreover, we are aware that the assumptions in the following list, from a mathematical point of view, are very similar. However, we believe that even if two assumptions may be equivalent mathematically, their different rephrasing may evoke different opinions and ideas. We hope that an engaged reader will consider some of these assumptions more reasonable than others and be surprised to discover how they are interconnected.
\begin{definition}[No Harm, No Foul]
Let $\boldsymbol{d},\boldsymbol{d}'\colon A\to \Delta^n$ be decisions on a  sensitive group $A$. Then
\[
    \boldsymbol{d}<_u \boldsymbol{d}' \Rightarrow \boldsymbol{d}<_f \boldsymbol{d}'  
\]
\end{definition}
The ``No Harm, No Foul'' assumption states that decisions can be unfair only when compared between different sensitive groups. In isolation, any decision regarding a single sensitive group is considered fair, which means that improving a decision for a single sensitive group can also be considered an improvement in fair probabilities. This resembles the point of view of earlier works on fairness, where fairness was considered simply as a constraint to satisfy~\cite{feldman2015certifying}.
\begin{definition}[Fairness as Optimization]
    Let $\boldsymbol{d},\boldsymbol{d}'\colon A\to \Delta^n$ be decisions on a  sensitive group $A$. Then
    \[
        \boldsymbol{d}<_f \boldsymbol{d}' \Rightarrow \boldsymbol{d}<_u \boldsymbol{d}'  
    \]
\end{definition}
The ``Fairness as Optimization'' assumption states that if, for a sensitive group, we theoretically manage to improve a decision according to fair probabilities, then we should at least be able to see the improvement for that group according to unfair probabilities as well. This means that an optimization step in fair probabilities can also be found in unfair probabilities. This, we believe, is the point of view of many in-processing methods~\cite{pmlrv28zemel13,zhang2018mitigating}.
\begin{definition}[Representation Matters]
    Let $\boldsymbol{d}\colon X\to \Delta^n$ be a decision. If  $\boldsymbol{d}\big|_{A}$ is not a maximal decision according to $\leq_u$ for some sensitive group $A$, then $\boldsymbol{d}$ is not unbiased.
\end{definition}
The ``Representation Matters'' assumption states that if we fail to correctly represent a sensitive group according to our data, then our decision cannot be considered fair. This is tied with the concept of ``cherry-picking'' in fairness literature \cite{dwork2012fairness,cherrypick_fleisher2021s}

\begin{definition}[Affirmative Action]
    Let $\boldsymbol{d}\colon X\to \Delta^n$ be an unbiased decision. There exist well-behaved sequences $(\boldsymbol{\omega}_{\Azero}^{(k)})_{k\in\NN}$, $(\boldsymbol{\omega}_{\Aone}^{(k)})_{k\in\NN}$ and associated decisions $\boldsymbol{d}_{\Azero}\colon \Azero\to \Delta^n$, $\boldsymbol{d}_{\Aone}\colon \Aone\to \Delta^n$ on $\leq_u$ such that $\boldsymbol{d}\big\vert_{\Azero}=\boldsymbol{d}_{\Azero}$ and $\boldsymbol{d}\big\vert_{\Aone}=\boldsymbol{d}_{\Aone}$.
\end{definition}
The ``Affirmative Action'' assumption states that in order to build a fair decision, we can set different thresholds for different sensitive groups to counterbalance the bias present in the data.

\begin{definition}[Double Standard]
    Let $\boldsymbol{d}\colon X\to \Delta^n$ a maximal decision for $\leq_u$. Then $\boldsymbol{d}\big\vert_{\Azero}$ and $\boldsymbol{d}\big\vert_{\Aone}$ are $\omega$-associated with two well-behaved 
    sequences $(\boldsymbol{\omega}_{\Azero}^{(k)})_{k\in\NN}$ and $(\boldsymbol{\omega}_{\Aone}'^{(k)})_{k\in\NN}$ according to $\leq_f$.
\end{definition}
The ``Double Standard'' assumption states that an optimal unfair decision treats the two communities as if there were two different ``standards'' for being selected as a candidate by the decision. This and the previous assumption are the two most common assumptions for post-processing methods~\cite{Corbett-Davies-faircost,menon-faircost,massaging1}. 
\section{Consequences of the Fairness Assumptions}
\subsection{Preliminary Results}
We start by showing how all the presented fairness assumptions imply either the ``Affirmative Action'' assumption or the ``Double Standard'' one. 
\begin{lemma}[No Harm, No foul $\Rightarrow$ Affirmative Action]
    \begin{proof}
        Suppose by contradiction, that for a sensitive group $A$ the decision $\boldsymbol{d}\big|_{A}$ cannot be associated with any well-behaved sequence according to $\leq_u$. Then, by Weller's Theorem, $\boldsymbol{d}\big|_{A}$ cannot be maximal on $\leq_u$. So there exists a decision $\boldsymbol{d}'\colon A\to \Delta^n$ such that $\boldsymbol{d}\big|_{A}<_u \boldsymbol{d}'$. By the No Harm, No Foul assumption we have $\boldsymbol{d}\big|_{A}<_f \boldsymbol{d}'$, then we can extend $\boldsymbol{d}'$ to a new decision on $X$ so defined
        \[
        	\boldsymbol{d}'(x)=
        	\begin{cases}
        		\boldsymbol{d}'(x) & \text{if } x\in A	\\	
        		\boldsymbol{d}(x) & \text{otherwise}
        	\end{cases}
        \]
        Clearly $\boldsymbol{d}<_f \boldsymbol{d}'$, contradicting the maximality of $\boldsymbol{d}$.
    \qed\end{proof}
\end{lemma}

\begin{lemma}[Fairness as Optimization $\Rightarrow$ Double Standard]
    \begin{proof}
        As before, with $p_f$ and $p_u$ swapped.
    \qed\end{proof}
\end{lemma}


\begin{lemma}[Representation Matters $\Rightarrow$ Affirmative Action]
    \begin{proof}
        Representation Matters can be restated as for an unbiased $\boldsymbol{d}$ then $\boldsymbol{d}\big\vert_A$ is maximal according to $\leq_u$. By Weller's Theorem we have the thesis.
    \qed\end{proof}
\end{lemma}

\subsection{The Main Consequence}
Since we have proven that our previous assumptions had to satisfy either Affirmative Action or Double Standard, we can now see how these final two assumptions require specific properties on the biasing process. Namely, that the biasing process or the debiasing process needs to be monotonic.
\begin{theorem}\label{teo:condition}
	The Affirmative Action assumption holds if and only if the probabilities satisfy the following conditions:
    \begin{enumerate}
    	\item\label{cond:1} for all $i\in [n]$ and for almost all $x\in X$ $$p_{u}(y_i\vert x)=0\Rightarrow p_{f}(y_i\vert x)=0$$
    	\item\label{cond:2} for all $i,j\in [n]$ and for almost all $x,x'\in A$
        \[
    	    \frac{p_{f}(y_i\vert x)}{p_{f}(y_j\vert x)}\leq\frac{p_{f}(y_i\vert x')}{p_{f}(y_j\vert x')}\Rightarrow \frac{p_{u}(y_i\vert x)}{p_{u}(y_j\vert x)}\leq\frac{p_{u}(y_i\vert x')}{p_{u}(y_j\vert x')}
    	\]
        for all sensitive groups $A\in\groups$.
    \end{enumerate}
    \begin{proof}
    	$(\Rightarrow\ref{cond:1})$ Suppose by contradiction that there exist $i$ and a non-zero set $S$ such that
    	\[
    	p_{f}(y_i\vert x)>0
    	\text{ and }
    	p_u(y_i\vert x)=0\quad \forall x\in S
    	\]
    	and let's consider an optimal decision $\boldsymbol{d}$ on the fair probabilities $w$-associated with $(\boldsymbol\omega^{(k)})_{k\in\NN}$ defined as follows
    	\[
    	\omega_s^{(k)}:=
    	\begin{cases}
    		1/k &\text{if } s = i\\
    		(k-1)/k(n-1) &\text{otherwise}
    	\end{cases}
    	\]
    	Clearly $d_i(S)=1$, let's now take a sensitive group $A$ such that $\mu(A\cap S)>0$. Such group must exists since the sensitive groups are a finite partition of $X$. Now we have $p_u(y_i\vert x)=0$ for all $x\in S\cap A$, so $\boldsymbol{d}\big|_{A}$ it is not optimal according to Theorem \ref{teo:weller}.\\
    	
        $(\Rightarrow \ref{cond:2})$ Let's suppose by way of contradiction that the implication is not true for some sensitive group $A$. Formally this means that there exists a non-zero subset $S\times S'\subseteq A\times A$ such that for all $x\in S, x'\in S'$ we have 
        \[
            \frac{p_{f}(y_i\vert x)}{p_{f}(y_j\vert x)}\leq\frac{p_{f}(y_i\vert x')}{p_{f}(y_j\vert x')}
        \quad\text{
        and}\quad         
            \frac{p_{u}(y_i\vert x)}{p_{u}(y_j\vert x)}>\frac{p_{u}(y_i\vert x')}{p_{u}(y_j\vert x')}
        \]
        Notice that we do not have to check the case where we have $p_{u}(y_i\vert x)/p_{u}(y_j\vert x)=0/0$ since we would also have  $p_{f}(y_i\vert x)/p_{f}(y_j\vert x)=0/0$ by condition \ref{cond:1}.
        Let's now take a threshold $t$ such that 
        \[
            \mu\left(\frac{p_{f}(y_i\vert x)}{p_{f}(y_j\vert x)}\leq t\leq \frac{p_{f}(y_i\vert x')}{p_{f}(y_j\vert x')} \bigg\vert\ (x,x')\in S\times S'\right)>0
        \]
        such threshold always exists since 
        \[
            \frac{p_{f}(y_i\vert x)}{p_{f}(y_j\vert x)}\leq\frac{p_{f}(y_i\vert x')}{p_{f}(y_j\vert x')}\quad\forall x\in S, x'\in S'
        \]
        implies
        \[
        \sup_{x\in S}\frac{p_{f}(y_i\vert x)}{p_{f}(y_j\vert x)}\leq\inf_{x'\in S'}\frac{p_{f}(y_i\vert x')}{p_{f}(y_j\vert x')}
        \]
		Consider now the following well-behaved sequence $(\boldsymbol\omega^{(k)})_{k\in\NN}$ defined as follows
		\[
		\omega_s^{(k)}:=
		\begin{cases}
			 t/k(t+1) &\text{if } s = i\\
			1/k(t+1) &\text{if } s = j\\
			(k-1)/k(n-2) &\text{otherwise}
		\end{cases}
		\]
		and let $\boldsymbol{d}$ be a decision $w$-associated with it.
		In particular we can take $\boldsymbol{d}$ such that if we consider the following sets
		\[
		S_{j}:=\left\{x\in S\colon \frac{p_{f}(y_i\vert x)}{p_{f}(y_j\vert x)}\leq t\right\}
		\]
		\[
		S'_{i}:=\left\{x'\in S'\colon \frac{p_{f}(y_i\vert x')}{p_{f}(y_j\vert x')}\geq t\right\}
		\]
		we have that $d_j(S_{j})>0$ and $d_i(S'_{i})>0$.\\
		We claim that such decision cannot be optimal according to $p_{u}$: if it were it would exist a $w$-associated well-behaved sequence $(\omega'^{(k)})_{k\in\NN}$. In particular the value $\omega_i^{(k)}/\omega_j^{(k)}$ converges to some value $t'$. Then for all $x\in S_j$ and $x'\in S'_i$
		\[
		t'\leq 
		\frac{p_{u}(y_i\vert x',a)}{p_{u}(y_j\vert x',a)}
		<
		\frac{p_{u}(y_i\vert x,a)}{p_{u}(y_j\vert x,a)}
		\leq
		t'
		\]
		which is absurd.\\
        \\
        $(\Leftarrow \ref{cond:1}+\ref{cond:2})$ We first prove that $\beta$, the biasing function that transforms fair probabilities to unfair ones does not depend on $x\in X$ almost everywhere. This follow directly from the hypothesis since almost everywhere we have
        \[
        \frac{p_{f}(y_i\vert x)}{p_{f}(y_j\vert x)}=\frac{p_{f}(y_i\vert x')}{p_{f}(y_j\vert x')}\Rightarrow \frac{p_{u}(y_i\vert x)}{p_{u}(y_j\vert x)}=\frac{p_{u}(y_i\vert x')}{p_{u}(y_j\vert x')}
        \]
        for all $i,j\in[n]$.\\
        Hence 
        \[
        p_{f}(y\vert x)=p_{f}(y\vert x')
        \Rightarrow p_{u}(y\vert x)=p_{u}(y\vert x')
        \]
        So people from the same sensitive group with the same fair probabilities are assigned the same unfair probabilities as well. So for all sensitive groups $A$, we can find a parameterized function $\beta^{A}$ such that almost everywhere
        \[
            p_{u}(y\vert x)=\beta\big(p_{f}(y\vert x),x\big) = \beta^{A}\big(p_{f}(y\vert x)\big)
        \]
        and for all $t,t'\in \Delta^n\cap (0,1)^n$
        \begin{equation}\label{property beta}
             \frac{t_i}{t_j}\leq\frac{t'_i}{t'_j} \Rightarrow \frac{\beta_i^{A}(t)}{\beta_j^{A}(t)}\leq\frac{\beta_i^{A}(t')}{\beta_j^{A}(t')}
         \end{equation}
        Let us now consider a optimal decision $\boldsymbol{d}$ according to fair probabilities $w$-associated with a well-behaved sequence $(\omega^{(k)})_{k\in\NN}$. Furthermore, we can assume that for all $i,j\in[n]$ the sequence $(\omega_i^{(k)}/\omega_j^{(k)})_{k\in\NN}$ is monotonic. If that was not the case for some $i,j$, that is if $(\omega_i^{(k)}/\omega_j^{(k)})_{k\in\NN}$ were not a monotonic sequence, we can consider a monotonic subsequence, which always exists since any real sequence has a monotonic subsequence. By repeating this process for all $i,j\in[n]$ we can ensure the sequence's monotonicity for every couple of components.
        We claim that for all sensitive groups $A$, the decision $\boldsymbol{d}\big|_{A}$ is, according to $p_u(y\vert x)$, $w$-associated with the well-behaved sequence $(\beta^{A}(\omega^{(k)}))_{k\in\NN}$ and hence maximal.\\
        First, let us show that the sequence is indeed well-behaved. Notice that because of condition \ref{cond:1} we know that $\beta^{A}(\Delta^n\cap (0,1)^n)\subseteq \Delta^n\cap (0,1)^n$ in particular $(\beta^{A}(\omega^{(k)}))_{k\in\NN}$ is indeed a sequence in $\Delta^n\cap (0,1)^n$. Also for all $i,j\in [n]$, the sequence $\left(\beta_i^{A}(\omega^{(k)})/\beta_j^{A}(\omega^{(k)})\right)_{k\in\NN}$ is monotonic and hence converges in $[0,+\infty]$. The monotonicity follows from the fact that
        \[
        		\frac{\omega_i^{(k)}}{\omega_j^{(k)}}\leq \frac{\omega_i^{(k+1)}}{\omega_j^{(k+1)}}
        		\Rightarrow
        		\frac{\beta_i^{A}(\omega^{(k)})}{\beta_j^{A}(\omega^{(k)})}\leq 	\frac{\beta_i^{A}(\omega^{(k+1)})}{\beta_j^{A}(\omega^{(k+1)})}
        \]
		Now we only need to prove that $\boldsymbol{d}\big|_{A}$ is $w$-associated with the sequence.
		We know that almost everywhere on $A$, for all $i,j\in [n]$
		\[
		d_i(x)>0 \Rightarrow \exists\ \overline{k}\in\NN\text{ s.t. }\frac{p_f(y_i\vert x)}{p_f(y_j\vert  x)}\geq \frac{\omega_i^{(k)}}{\omega_j^{(k)}} \quad \forall k\geq \overline{k}
		\]
		then using property \ref{property beta} for $\beta^{A}$ we obtain for all $k\geq \overline{k}$
		\[
		\frac{p_u(y_i\vert x)}{p_u(y_j\vert  x)}=\frac{\beta^{A}(p_f(y_i\vert x))}{\beta^{A}(p_f(y_j\vert  x))}\geq \frac{\beta_i^{A}(\omega^{(k)})}{\beta_j^{A}(\omega^{(k)})}
		\]
		Hence
		\[
		d_i(x)>0 \Rightarrow \exists\ \overline{k}\in\NN\text{ s.t. }
		\frac{p_u(y_i\vert x)}{p_u(y_j\vert  x)}\geq \frac{\beta_i^{A}(\omega^{(k)})}{\beta_j^{A}(\omega^{(k)})} \quad \forall k\geq \overline{k}
		\]
    \qed\end{proof} 
\end{theorem}

\begin{theorem}
    The Double Standard assumption holds if and only if the probabilities satisfy the following conditions:
    \begin{enumerate}
    	\item for all $i\in [n]$ and for almost all $x\in X$ $$p_{f}(y_i\vert x)=0\Rightarrow p_{u}(y_i\vert x)=0$$
    	\item for all $i,j\in [n]$ and for almost all $x,x'\in A$\[
    	\frac{p_{u}(y_i\vert x)}{p_{u}(y_j\vert x)}\leq\frac{p_{u}(y_i\vert x')}{p_{u}(y_j\vert x')}\Rightarrow \frac{p_{f}(y_i\vert x)}{p_{f}(y_j\vert x)}\leq\frac{p_{f}(y_i\vert x')}{p_{f}(y_j\vert x')}
    	\]
        for all sensitive groups $A\in\groups$.
    \end{enumerate}
    \begin{proof}
    	The proof is analogous to the one of Theorem \ref{teo:condition} with $p_u$ and $p_f$ swapped.\\
    \qed\end{proof}
\end{theorem}
In the binary case $Y=\{0,1\}$, the requirements on the biasing (and debiasing) function $\beta$ can be simplified as follows:
\begin{corollary}
    Let $Y=\{0,1\}$ be a binary label. The Affirmative Action assumption holds if and only if, for almost all $x,x'\in A$, it holds
    \begin{enumerate}
    	\item $p_{u}(y=1\vert x)=0\Rightarrow p_{f}(y=1\vert x)=0$
    	\item $p_{u}(y=1\vert x)=1\Rightarrow p_{f}(y=1\vert x)=1$
    	\item $p_{f}(y=1\vert x)\leq p_{f}(y=1\vert x')\Rightarrow p_{u}(y=1\vert x)\leq p_{u}(y=1\vert x')$
    \end{enumerate}
    where $A$ is a sensitive group. In particular, on each sensitive group, the biasing function $\beta$ is non-decreasing function of $p_{f}(y=1\vert x)$.
    \begin{proof}
        It follows directly from Theorem \ref{teo:condition} and the fact, because the label is binary, we have $p(y=0\vert x)+p(y=\vert x)=1$.
    \qed\end{proof}
\end{corollary}
And, similarly, for the Double Standard assumption:
\begin{corollary}
    Let $Y=\{0,1\}$ be a binary label. The Double Standard assumption holds if and only if, for almost all $x,x'\in A$, it holds
    \begin{enumerate}
    	\item $p_{f}(y=1\vert x)=0\Rightarrow p_{u}(y=1\vert x)=0$
    	\item $p_{f}(y=1\vert x)=1\Rightarrow p_{u}(y=1\vert x)=1$
    	\item $p_{u}(y=1\vert x)\leq p_{u}(y=1\vert x')\Rightarrow p_{f}(y=1\vert x)\leq p_{f}(y=1\vert x')$
    \end{enumerate}
    where $A$ is a sensitive group. In particular, on each sensitive group, the debiasing function $\beta$ is non-decreasing function of $p_{u}(y=1\vert x)$.
\end{corollary}
As previously stated, for a binary label, the assumptions we have shown are requiring that the biasing or the debiasing process is monotonic. When the label is not binary, the requirements are more complex, but the idea remains the same: the process must be relatively monotonic according to the ratio between labels. It's also noteworthy that, unless bias is strictly monotonic, the Affirmative Action and Double Standard assumptions are not exactly equivalent. 
\section{Discussion}
In the previous section, we have proven how all the assumptions we have presented imply that the biasing process behaves as a monotonic function. Given this result, it would be interesting to see if fair models actually behave accordingly. Luckily, there is already some literature that has started to investigate this.
In their paper ``Reranking individuals: The effect of fair classification within-groups''\cite{goethals2024beyond}, Goethals et al. have researched how fair machine learning models behave with respect to each sensitive group. In particular, they measure how often different fair models change the ranking of individuals within a sensitive group. They first adopt a fair-agnostic model to create a baseline ranking based on the score of each individual. Then, they apply different fair models to the same data and measure how often the ranking changes. To do so, they use the Kendall Tau coefficient, a measure of the similarity between two rankings, defined as
\[
    \frac{\text{number of concordant pairs}-\text{number of discordant pairs}}{\text{number of pairs}}
\]
Based on their results, they conclude that a number of models often inexplicably rerank individuals within the same group. In particular, they caution against the use of pre- and in-processing methods when no within-group bias is expected, since these methods are the ones that suffer the most from this issue.

Our theoretical framework is necessary to formally support their claims. They correctly point out that the reranking of individuals within a group is a sign that the model is not behaving as expected, and they intuitively connect this to the fact that this behavior can be considered acceptable only when bias affects even people within the same sensitive group.

Given our results, we can now formally claim that if the within-group reranking happens, it means that all the assumptions listed in Section \ref{sec:assumptions} are to be considered false.

For some models, this is worrisome: a number of them, in fact, seem to operate under the "Fairness as Optimization" assumption. This means that the reranking of individuals should not be an expected outcome of these models.

In other words, the models seem to be self-contradicting: they operate under the assumption that being as faithful as possible to the data is important, yet they fail to be optimal according to their own assumptions.

Another important discussion regards some of the claims from \cite{menon-faircost,Corbett-Davies-faircost}, where they prove that the best model can be found by applying different thresholds on the probability distribution for each sensitive group. Our paper shows how "being best" in this context is equivalent to believing that the biasing process is monotonic. And since the process is monotonic, then indeed the best model can be found by applying thresholds, as a consequence of Weller's Theorem.

This conclusion is tautological; these papers claim that their models are ``the best" because they follow their definition of what "best" is.
\section{Conclusions}
In this paper, we have shown how assumptions regarding fair models are tied to assumptions about the probabilities of the data. We believe that being transparent about these assumptions is crucial to understanding the behavior of a model.

As others have already claimed, in order to make a fair model, practitioners should start by understanding how bias influences the data.

One of the first questions we need to ask ourselves is whether bias is monotonic or not. This is important, especially when discrimination is possible even among people of the same sensitive group.
Two possible outcomes arise from asking how bias behaves:
\begin{itemize}
    \item Bias is monotonic. Then, searching for the best model is equivalent to searching for the best thresholds. However, it is also important to note that the fairest model is not necessarily the one that satisfies a fairness constraint. As proven in \cite{favier2023fair}, even when bias is monotonic, depending on the kind of bias, the fairest model can be very unintuitive.
    \item Bias is not monotonic. If this is the case, we need to understand why it is not so and determine if we can identify the people who are more likely to be discriminated against. If bias is based on different demographic characteristics, it's likely that it affects people who correlate with stereotypes from other groups. This begs the question: should fairness be based on how predictable the sensitive attribute is?
\end{itemize}
The answers to these questions cannot be universal but need to be weighed based on the problem at hand. Often, educated assumptions must be made in order to build a fair model. Despite what some old proverbs may say, assuming is not wrong, as long as we are explicit about it.
\bibliographystyle{splncs04}
\bibliography{biblio.bib}
\end{document}